\documentclass[sigconf]{acmart}
\AtBeginDocument{%
  }
\usepackage{enumitem}
\setlist[itemize]{leftmargin=*}
\setcopyright{acmlicensed}
\copyrightyear{2018}
\acmYear{2018}
\acmDOI{XXXXXXX.XXXXXXX}
\acmConference[Conference acronym 'XX]{Make sure to enter the correct
  conference title from your rights confirmation email}{June 03--05,
  2018}{Woodstock, NY}
\acmISBN{978-1-4503-XXXX-X/2018/06}

\begin{document}

%%
%% The "title" command has an optional parameter,
%% allowing the author to define a "short title" to be used in page headers.
\title{SonarLLM: A Native Sonar--Optical Multimodal Large Language Model for Underwater Perception}

%%
%% The "author" command and its associated commands are used to define
%% the authors and their affiliations.
%% Of note is the shared affiliation of the first two authors, and the
%% "authornote" and "authornotemark" commands
%% used to denote shared contribution to the research.
\author{Cong Su}
\email{siliconevolution@gmail.com}
\affiliation{%
  \institution{Faculty of Information Engineering and Automation, Kunming University of Science and Technology}
  \city{Kunming}
  \country{China}}
\affiliation{%
  \institution{Yunnan Key Laboratory of Artificial Intelligence}
  \city{Kunming}
  \country{China}}

\author{Longxuan Ma}
\authornote{Corresponding author.}
\email{lxma@kust.edu.cn}
\affiliation{%
  \institution{Faculty of Information Engineering and Automation, Kunming University of Science and Technology}
  \city{Kunming}
  \country{China}}
\affiliation{%
  \institution{Yunnan Key Laboratory of Artificial Intelligence}
  \city{Kunming}
  \country{China}}

\author{Ling Dong}
\email{ling.dong@kust.edu.cn}
\affiliation{%
  \institution{Faculty of Information Engineering and Automation, Kunming University of Science and Technology}
  \city{Kunming}
  \country{China}}
\affiliation{%
  \institution{Yunnan Key Laboratory of Artificial Intelligence}
  \city{Kunming}
  \country{China}}

\author{Guofeng Tang}
\email{920903629@qq.com}
\affiliation{%
  \institution{Faculty of Information Engineering and Automation, Kunming University of Science and Technology}
  \city{Kunming}
  \country{China}}
\affiliation{%
  \institution{Yunnan Key Laboratory of Artificial Intelligence}
  \city{Kunming}
  \country{China}}

\author{Weijie Yin}
\email{18102382274@163.com}
\affiliation{%
  \institution{Faculty of Information Engineering and Automation, Kunming University of Science and Technology}
  \city{Kunming}
  \country{China}}
\affiliation{%
  \institution{Yunnan Key Laboratory of Artificial Intelligence}
  \city{Kunming}
  \country{China}}

\author{Haohui Chen}
\email{2954390791@qq.com}
\affiliation{%
  \institution{Faculty of Information Engineering and Automation, Kunming University of Science and Technology}
  \city{Kunming}
  \country{China}}
\affiliation{%
  \institution{Yunnan Key Laboratory of Artificial Intelligence}
  \city{Kunming}
  \country{China}}

\author{Zhengtao Yu}
\email{ztyu@hotmail.com}
\affiliation{%
  \institution{Faculty of Information Engineering and Automation, Kunming University of Science and Technology}
  \city{Kunming}
  \country{China}}
\affiliation{%
  \institution{Yunnan Key Laboratory of Artificial Intelligence}
  \city{Kunming}
  \country{China}}

\renewcommand{\shortauthors}{Su et al.}

%%
%% The abstract is a short summary of the work to be presented in the
%% article.
\begin{abstract}
Reliable underwater perception requires complementary sensing under variable visibility. Optical cameras capture appearance and semantics but degrade rapidly with turbidity, whereas imaging sonar preserves geometry while exhibiting distinct range–azimuth structure and acoustic artifacts. Existing MLLMs, built primarily on optical encoders, are therefore ill-suited to model sonar or adaptively exploit sonar–optical complementarity. We propose SonarLLM, a sonar–optical MLLM that treats sonar as a native perceptual modality. It combines a sonar-specific encoder, modality-specific physics-aware feature enhancement, and reliability-aware hierarchical fusion to align acoustic structure with optical semantics and dynamically adjust their contributions as sensing quality changes. We also introduce SonarBench, a paired benchmark that spans four tasks---recognition, counting, visual question answering, and captioning---and, across the benchmark, three input settings: sonar-only, optical-only, and fusion. By fixing the scene and sonar observation while varying optical degradation, SonarBench enables controlled measurement of cross-modal complementarity. SonarLLM achieves 72.0\% macro accuracy across sonar-only recognition, counting, and VQA, outperforming the strongest baseline by 34.4 percentage points, and 68.7\% under fusion, exceeding the best baseline by 25.1 points. For recognition and counting, the fusion-over-optical gain grows from 6.0 to 36.0 points as turbidity increases, indicating the increasing complementary value of sonar under controlled optical degradation. Together, these results show that robust heterogeneous perception depends not only on adding sonar, but on representing and weighting it according to its sensing characteristics.

\end{abstract}

%%
%% The code below is generated by the tool at http://dl.acm.org/ccs.cfm.
%% Please copy and paste the code instead of the example below.
%%



%%
%% Keywords. The author(s) should pick words that accurately describe
%% the work being presented. Separate the keywords with commas.
%% A "teaser" image appears between the author and affiliation
%% information and the body of the document, and typically spans the
%% page.

\received{20 February 2007}
\received[revised]{12 March 2009}
\received[accepted]{5 June 2009}

%%
%% This command processes the author and affiliation and title
%% information and builds the first part of the formatted document.
\maketitle

\section{Introduction}

\begin{figure*}[t]
	\centering
	\includegraphics[width=0.98\textwidth]{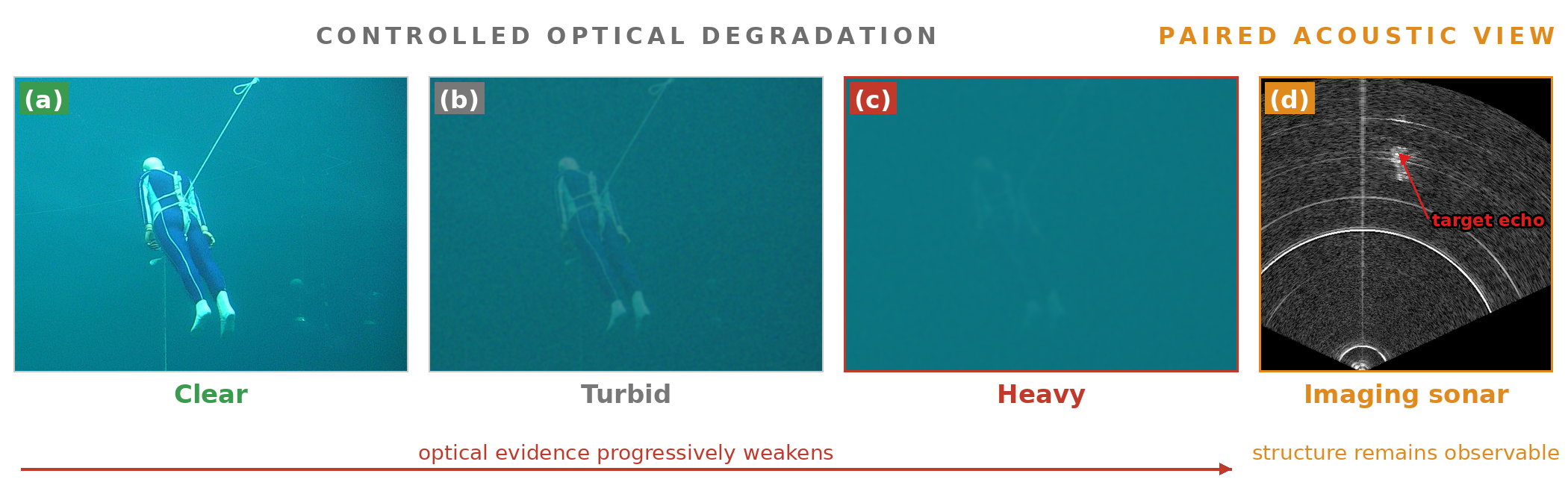}
	\caption{Paired sonar--optical observations under controlled optical degradation. Optical evidence weakens with turbidity, whereas sonar preserves structural cues for the same scene.}
	\label{fig:teaser}
	\vspace{-5pt}
\end{figure*}

Underwater environments impose severe and highly variable sensing conditions. Recent multimodal large language models (MLLMs) have advanced open-ended visual understanding, while underwater models such as MarineGPT~\citep{zheng2023marinegpt} and NAUTILUS~\citep{xu2025nautilus} extend these capabilities to marine image captioning, question answering, and recognition. However, these models remain predominantly dependent on optical imagery. Absorption and scattering progressively erase color, texture, contrast, and object boundaries as turbidity increases~\citep{liu2020uieb}. Imaging sonar is largely independent of illumination and can preserve object contours, spatial structure, and range information under poor visibility~\citep{neupane2020sonarreview}. Optical and sonar observations are therefore complementary: one provides appearance and semantic detail, whereas the other supplies more stable geometric evidence under optical degradation~\citep{li2025rgbs50}, as illustrated in Fig.~\ref{fig:teaser}.

Exploiting this complementarity requires more than adding sonar as another visual input. Existing sonar--optical systems are primarily designed for task-specific detection or tracking~\citep{li2025rgbs50,wu2026umod} and do not address open-ended language reasoning. Natural-image encoders are also poorly matched to sonar, whose range--azimuth geometry and artifacts include speckle, reverberation, acoustic shadows, and range-dependent propagation loss~\citep{neupane2020sonarreview,steiniger2022sonarsurvey}. Moreover, modality reliability is observation-dependent: optical evidence may collapse under turbidity, while sonar can be corrupted by acoustic noise and artifacts, making fixed fusion vulnerable to an unreliable sensor~\citep{park2025resilient}. Together, these limitations motivate a sonar--optical MLLM that jointly addresses modality-specific representation, degradation-aware enhancement, and reliability-aware interaction across sensors and semantic levels.

To address these key challenges, we propose \textbf{SonarLLM}, a sonar--optical MLLM that treats imaging sonar as a native perceptual modality rather than an auxiliary image. First, for modality-specific representation, SonarLLM retains the pretrained Qwen3-VL-8B optical encoder~\citep{bai2025qwen3vl} and introduces an independent sonar encoder equipped with a multi-scale Sonar Stem and range--azimuth positional encoding. Second, for modality-dependent degradation, dedicated Visual Feature Enhancement (VFE) modules operate in each modality: Optical-VFE targets scattering-related corruption, whereas Acoustic-VFE targets reverberation-related components and range attenuation. Third, AGFM predicts quality-aware modality weights, while dual-stream hierarchical DeepStack delivers reweighted optical and sonar features to multiple language-model layers. Finally, progressive training establishes sonar-domain representations, cross-modal correspondence, reliability learning, and instruction-following capability.

We further introduce \textbf{SonarBench}, a paired benchmark for controlled evaluation of sonar--optical understanding. It covers recognition, counting, visual question answering, and captioning and spans sonar-only, optical-only, and fusion settings across the benchmark as a whole. Fusion evaluation focuses on the three accuracy-based tasks, while captioning provides a separate diagnostic of open-ended generation. Optical observations are evaluated from clear to heavily turbid conditions. Unlike protocols that compare different samples across conditions, SonarBench fixes the underlying scene and sonar observation while varying only optical quality. This design separates gains from improved unimodal modeling from gains attributable to complementary sonar evidence as optical reliability deteriorates.

Experiments demonstrate both strong sonar understanding and robust cross-modal complementarity. SonarLLM achieves \textbf{72.0\%} macro accuracy across sonar-only recognition, counting, and VQA, outperforming the strongest baseline by \textbf{34.4 percentage points}; substantially larger 27B--35B general-purpose MLLMs do not close this gap. Under fusion input, SonarLLM reaches \textbf{68.7\%}, exceeding the best baseline by \textbf{25.1 points}. For recognition and counting, the fusion-over-optical gain increases from \textbf{6.0 points} under clear conditions to \textbf{36.0 points} under heavy turbidity, while fusion performance remains comparatively stable. SonarLLM achieves the best overall performance among the evaluated MLLMs on SonarBench.

Our main contributions are summarized as follows:
\begin{itemize}
	\item We formulate sonar--optical language understanding as a heterogeneous sensing problem and propose \textbf{SonarLLM}, which unifies sonar-specific representation, modality-specific feature enhancement, reliability-aware gating, and hierarchical interaction within a shared language model.
	
	\item We introduce \textbf{SonarBench}, a paired benchmark whose controlled intervention keeps the scene and sonar observation fixed while varying optical quality, thereby isolating unimodal sonar capability from cross-modal complementarity.
	
	\item Through controlled comparisons, representation analysis, and structural ablations, we show that native sonar modeling provides capabilities not recovered by model scale or instruction tuning alone, while reliability-aware fusion becomes increasingly valuable as optical evidence deteriorates.
\end{itemize}

\section{Related Work}

\subsection{Underwater Vision-Language and Sonar Perception}

Recent work has extended vision-language learning to marine and underwater environments. MarineGPT~\citep{zheng2023marinegpt} uses domain-specific image--text data and instruction tuning for captioning, question answering, and recognition, while AquaticCLIP~\citep{alawode2025aquaticclip} learns underwater vision--language representations through large-scale contrastive pretraining. NAUTILUS~\citep{xu2025nautilus} incorporates physics-aware enhancement, and OceanGPT~\citep{bi2024oceangpt} targets broader ocean-domain knowledge. OceanPile~\citep{xue2026oceanpile} and OceanGym~\citep{xue2025oceangym} provide ocean multimodal data and embodied evaluation, respectively, but do not address sonar-native open-ended reasoning under paired optical degradation. These efforts demonstrate the value of domain data and priors, yet do not study the heterogeneous language-level interaction considered here.

Imaging sonar provides complementary structural observations independent of ambient illumination. Existing sonar research primarily addresses detection, classification, segmentation, and tracking~\citep{neupane2020sonarreview,steiniger2022sonarsurvey}, supported by datasets such as UATD~\citep{xie2022uatd} and SCTD~\citep{zhang2022sctd}. Paired datasets including RGBS50~\citep{li2025rgbs50}, UMOD~\citep{wu2026umod}, and SOVIS~\citep{chen2026sovis} further demonstrate the value of sonar--optical fusion for specific discriminative tasks. However, these systems typically fuse task-dependent features or predictions and do not connect acoustic representations to open-ended language reasoning. SonarLLM instead models sonar as a native perceptual modality and enables continuous interaction among acoustic structure, optical semantics, and language representations.

\subsection{Heterogeneous Multimodal Fusion and Evaluation}

General-purpose MLLMs employ diverse mechanisms to connect visual and linguistic representations. LLaVA~\citep{liu2023visualinstruction,liu2023llava15} uses learned projection, BLIP-2~\citep{li2023blip2} introduces a Q-Former, Flamingo~\citep{alayrac2022flamingo} injects visual context through cross-layer attention, and Qwen3-VL~\citep{bai2025qwen3vl} incorporates hierarchical visual features through DeepStack. Gated and reliability-aware fusion methods~\citep{arevalo2017gated,park2025resilient} additionally adjust modality contributions according to sensor quality. Nevertheless, these approaches generally assume visually homogeneous inputs or do not explicitly account for sensors with different imaging geometry, degradation processes, and environment-dependent reliability. SonarLLM combines modality-specific representation and enhancement with reliability-aware hierarchical interaction to address these requirements jointly.

Existing underwater evaluation resources are similarly divided by modality or task. UIEB~\citep{liu2020uieb} and EUVP~\citep{islam2020euvp} focus on optical enhancement; UATD and SCTD support sonar perception; and RGBS50, UMOD, and SOVIS target specific paired-sensor tasks. Vision-language benchmarks such as NautData~\citep{xu2025nautilus} and UWBench~\citep{zhang2025uwbench} remain centered on optical observations. They therefore cannot isolate whether multimodal gains arise from stronger unimodal modeling or from genuinely complementary sonar evidence. SonarBench addresses this limitation through paired interventions that keep the scene and sonar observation fixed while varying only optical quality, enabling controlled evaluation of sonar representation, cross-modal complementarity, and reliability adaptation.

\section{Method}
\label{sec:method}

\subsection{Overall Architecture}
\label{sec:overview}

\begin{figure*}[t]
	\centering
	\includegraphics[
	width=0.98\textwidth,
	keepaspectratio
	]{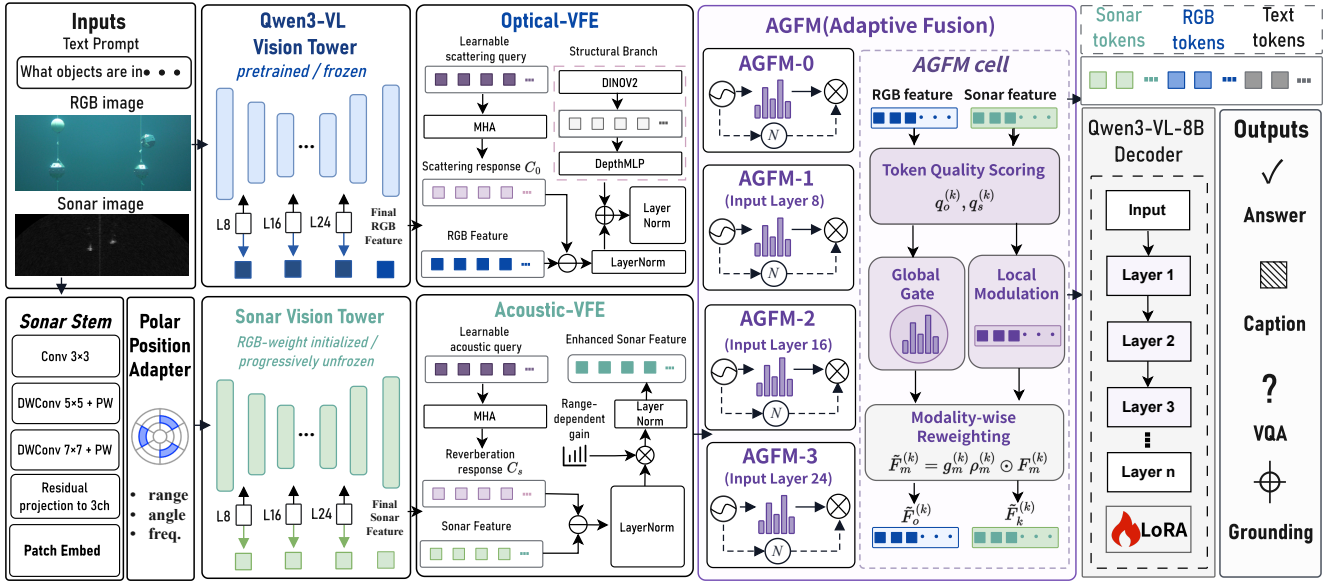}
	\caption{Overall architecture of SonarLLM, comprising sonar-native representation, modality-specific feature enhancement, reliability-aware hierarchical fusion, and progressive training.}
	\label{fig:architecture}
	\vspace{-5pt}
\end{figure*}

Given an optical image $I_o$, an imaging-sonar observation $I_s$, and a textual instruction $x$, SonarLLM extracts heterogeneous visual representations and performs joint reasoning within a shared language model. As shown in Fig.~\ref{fig:architecture}, it preserves the pretrained optical encoder $\mathcal E_o$ of Qwen3-VL-8B~\citep{bai2025qwen3vl} and introduces an independent sonar encoder $\mathcal E_s$, thereby treating sonar as a native perceptual modality rather than an auxiliary image input.

Each visual encoder produces a final feature $F_m$ and three intermediate features $\{F_m^{(k)}\}_{k=1}^{3}$ extracted after Transformer blocks 8, 16, and 24, where $m\in\{o,s\}$. Final features are processed by modality-specific Visual Feature Enhancement (VFE) modules, projected into the language space, and reweighted by AGFM$_0$. Intermediate features bypass the VFEs and are processed by AGFM$_k$ before being projected and injected into language-model layers 8, 16, and 24, respectively, through dual-stream DeepStack. This separation preserves modality-specific information while enabling cross-modal interaction at multiple semantic depths.

\subsection{Sonar-Native Visual Representation}
\label{sec:sonar_encoder}

Natural-image encoders are poorly matched to sonar statistics and range--azimuth geometry. We initialize a Polar-aware Sonar Vision Transformer (PSVT) from the Qwen3-VL visual tower to retain transferable visual priors, and adapt it through a multi-scale Sonar Stem and explicit geometric positional encoding.

\paragraph{Multi-scale Sonar Stem.}
Before patch embedding, the Sonar Stem captures echoes, boundaries, and acoustic shadows at different receptive fields:
\begin{equation}
	\begin{aligned}
		\Delta I_s
		&=
		\mathrm{Conv}_{1\times1}
		\left[
		f_7\left(f_5\left(f_3(I_s)\right)\right)
		\right],\\
		\widetilde I_s
		&=
		I_s+\gamma\Delta I_s ,
	\end{aligned}
	\label{eq:sonar_stem}
\end{equation}
where $f_3$, $f_5$, and $f_7$ denote convolutional transformations with different receptive fields. The zero-initialized coefficient $\gamma$ gradually introduces sonar-specific structure without disrupting the transferred representation at initialization.

\paragraph{Range--Azimuth Positional Adaptation.}
For a sonar patch centered at range $r_i$ and azimuth $\theta_j$, we augment the transferred positional representation as:
\begin{equation}
	z_{ij}^{s}
	=
	z_{ij}^{\mathrm{base}}
	+
	\lambda_p
	\left[
	\phi_r(r_i)+\phi_\theta(\theta_j)
	\right],
	\label{eq:polar_pe}
\end{equation}
where $\phi_r$ and $\phi_\theta$ encode physical range and azimuth, respectively. PSVT can therefore retain pretrained visual priors while explicitly adapting to sonar imaging geometry.

\subsection{Modality-Specific Feature Enhancement}
\label{sec:vfe}

Optical and sonar observations undergo different physical degradations. A
compact abstraction of their image-formation processes is:
\begin{equation}
	\begin{aligned}
		I_o(u) &= J(u)e^{-\beta d(u)}
		+B_\infty\left[1-e^{-\beta d(u)}\right],\\
		S(r,\theta) &= \frac{S_0TS(\theta)}{r^2}e^{-2\alpha_{\mathrm{phy}} r}
		+R(r,\theta),
	\end{aligned}
	\label{eq:physical_models}
\end{equation}
where $d(u)$ is scene range, $\beta$ is optical attenuation, $J(u)$ is
undegraded scene radiance, and $B_\infty$ is backscattered light. In the
sonar model, $S_0$ is a source-level constant, $\alpha_{\mathrm{phy}}$ is acoustic
attenuation, $TS(\theta)$ is target strength, and $R(r,\theta)$ is
reverberation.
Motivated by these distinct attenuation and interference terms, we apply
separate Optical-VFE and Acoustic-VFE modules to the final high-level
features, while intermediate DeepStack features bypass the VFEs. The modules
perform feature-space correction rather than inversion of the raw physical
image-formation processes.

\paragraph{Optical-VFE.}
A learnable query $e_b$ estimates feature corruption related to scattering in
$F_o$, while a frozen DINOv2-L encoder~\citep{oquab2023dinov2}
provides a generic structural reference $D_o$:
\begin{equation}
	\begin{aligned}
		C_o
		&=
		\mathrm{MHA}(e_b,F_o,F_o),\\
		F_o^{c}
		&=
		\mathrm{LN}\left(
		F_o-W_bC_o
		\right),\\
		\overline F_o
		&=
		\mathrm{LN}\left[
		F_o^{c}+\Psi_o(D_o)
		\right].
	\end{aligned}
	\label{eq:optical_vfe}
\end{equation}
Here, $W_b$ is a learned channel projection, and $\Psi_o$ maps $D_o$ to the
token shape of $F_o^c$. Optical-VFE performs feature correction rather than
physical inversion.

\paragraph{Acoustic-VFE.}
A learnable acoustic query $e_r$ estimates a feature component associated with
reverberation, while token range parameterizes learned compensation for
range-dependent attenuation:
\begin{equation}
	\begin{aligned}
		C_s
		&=
		\mathrm{MHA}(e_r,F_s,F_s),\\
		F_s^{c}[i]
		&=
		F_s[i]-\sigma(g_r)W_sC_s,\\
		G(r_i)
		&=
		\exp\!\left[
		2\log\left(
		\frac{r_i}{r_{\min}}
		\right)
		+2\alpha_c(r_i-r_{\min})
		\right],\\
		\overline F_s[i]
		&=
		\mathrm{LN}\left(
		G(r_i)F_s^{c}[i]
		\right).
	\end{aligned}
	\label{eq:acoustic_vfe}
\end{equation}
Here, $W_s$ is a learned channel projection, $\sigma$ is sigmoid, and $g_r$ is
a scalar gate. The learned range-compensation coefficient
$\alpha_c=\mathrm{softplus}(\widehat{\alpha}_c)$ is distinct from the physical
$\alpha_{\mathrm{phy}}$ in Eq.~\eqref{eq:physical_models}.
Because $r_i$ is the physical range of
the sonar token, the compensation retains the native range--azimuth geometry.

\subsection{Reliability-Aware Hierarchical Fusion}
\label{sec:agfm}
\label{sec:deepstack}

The reliability of optical and sonar observations varies across both environments and spatial regions. SonarLLM therefore combines the Adaptive Gated Fusion Module (AGFM) with dual-stream DeepStack to model global sensor reliability and local token quality at multiple visual levels.

At level $k$, let
$F_m^{(k)}=\{f_{m,i}^{(k)}\}_{i=1}^{N_m}$.
For $k=0$, $F_m^{(0)}$ denotes the enhanced and projected final representation; for $k\in\{1,2,3\}$, it denotes the feature extracted after visual block 8, 16, or 24. AGFM first estimates token quality and aggregates it into modality-level reliability:
\begin{equation}
	\begin{aligned}
		q_{m,i}^{(k)}
		&=
		h_m^{(k)}
		\left(
		\mathrm{LN}
		\left(
		f_{m,i}^{(k)}
		\right)
		\right),
		\qquad
		\bar q_m^{(k)}
		=
		N_m^{-1}
		\sum_{j=1}^{N_m}
		q_{m,j}^{(k)},\\
		\overline{\mathbf q}^{(k)}
		&=
		\left[
		\bar q_o^{(k)},
		\bar q_s^{(k)}
		\right],
		\qquad
		\mathbf g^{(k)}
		=
		\mathrm{softmax}
		\left(
		\overline{\mathbf q}^{(k)}/\tau_k
		\right).
	\end{aligned}
	\label{eq:reliability_gate}
\end{equation}
Here, $h_m^{(k)}:\mathbb R^{d_k}\!\to\!\mathbb R$ is a token-shared scalar
scorer. The vector $\mathbf g^{(k)}=[g_o^{(k)},g_s^{(k)}]$ contains relative
modality weights used as controlled reliability proxies. Each $\tau_k$ is
learnable, lower-bounded at $0.05$, and initialized to $2.0$.

AGFM further captures spatially non-uniform quality through mean-preserving token modulation:
\begin{equation}
	\begin{aligned}
		u_{m,i}^{(k)}
		&=
		\frac{
			1+\sigma\left(q_{m,i}^{(k)}\right)
		}{2},
		\qquad
		\bar u_m^{(k)}
		=
		N_m^{-1}
		\sum_{j=1}^{N_m}
		u_{m,j}^{(k)},\\
		\rho_{m,i}^{(k)}
		&=
		\frac{
			u_{m,i}^{(k)}
		}{
			\bar u_m^{(k)}
		},
		\qquad
		\widetilde f_{m,i}^{(k)}
		=
		g_m^{(k)}
		\rho_{m,i}^{(k)}
		f_{m,i}^{(k)}.
	\end{aligned}
	\label{eq:local_gate}
\end{equation}
The global factor $g_m^{(k)}$ allocates reliability across sensors, while $\rho_{m,i}^{(k)}$ redistributes importance among tokens without changing their mean scale. For single-modality input, AGFM reduces to an identity mapping.

Crucially, AGFM reweights rather than merges the two modality sequences. At the final level, the reweighted optical and sonar tokens form the initial multimodal context. At the three intermediate levels, the reweighted features are projected by modality-specific DeepStack mergers and injected into language layers 8, 16, and 24, respectively:
%\begin{equation}
%	\begin{aligned}
	%		H^{(\ell_k)}
	%		\leftarrow
	%		H^{(\ell_k)}
	%		&+
	%		\mathcal J_o^{(k)}
	%		\left[
	%		\mathcal P_o^{(k)}
	%		\left(
	%		\widetilde F_o^{(k)}
	%		\right)
	%		\right]\
	%		&+
	%		\mathcal J_s^{(k)}
	%		\left[
	%		\mathcal P_s^{(k)}
	%		\left(
	%		\widetilde F_s^{(k)}
	%		\right)
	%		\right].
	%	\end{aligned}
%	\label{eq:deepstack_injection}
%\end{equation}
\begin{equation}
	\begin{aligned}
		H^{(\ell_k)}
		&\leftarrow
		H^{(\ell_k)}
		+
		\mathcal J_o^{(k)}
		\left[
		\mathcal P_o^{(k)}
		\left(
		\widetilde F_o^{(k)}
		\right)
		\right]\\
		&\quad+
		\mathcal J_s^{(k)}
		\left[
		\mathcal P_s^{(k)}
		\left(
		\widetilde F_s^{(k)}
		\right)
		\right].
	\end{aligned}
	\label{eq:deepstack_injection}
\end{equation}
Equation~\eqref{eq:deepstack_injection} is applied at language layers 8, 16,
and 24. Here, $\mathcal P_m^{(k)}$ is a modality-specific DeepStack merger that
maps intermediate features to the language-model hidden space, and
$\mathcal J_m^{(k)}$ aligns and adds them to the visual-token span of the
corresponding modality. Maintaining separate optical and sonar streams before
injection avoids premature compression of their heterogeneous representations.

\subsection{Progressive Training Strategy}
\label{sec:training}

We train SonarLLM in four stages that progressively establish sonar-domain representations, acoustic semantics, cross-modal reliability, and language-level reasoning. Direct joint optimization would require limited paired data to simultaneously resolve domain shift, semantic organization, sensor correspondence, gate calibration, and instruction following. The staged curriculum first stabilizes acoustic representations and then introduces cross-modal and language supervision, reducing interference among these heterogeneous objectives. Table~\ref{tab:training_schedule} summarizes the resulting schedule.

\begin{table}[t]
	\centering
	\caption{Progressive training schedule. CE denotes category cross-entropy.}
	\label{tab:training_schedule}
	\small
	\setlength{\tabcolsep}{2.8pt}
	\begin{tabular}{@{}cll@{}}
		\toprule
		Stage & Data/objective & Trainable modules \\
		\midrule
		I & Unlabeled sonar / MAE & PSVT, decoder \\
		II & Labeled sonar / CE & PSVT/Stem, classifier \\
		III & Paired / $\mathcal L_{\mathrm{align}}$ & Sonar path, VFEs, AGFM \\
		IV & Instructions / $\mathcal L_{\mathrm{inst}}$ & LoRA, interfaces \\
		\bottomrule
	\end{tabular}
	\vspace{-15pt}
\end{table}

\paragraph{Stage I: Sonar-Domain Adaptation.}
PSVT is first adapted using masked reconstruction on unlabeled sonar images. Because low-response background occupies a large portion of each frame, normalized masked patches $\bar x_i$ are weighted by their local variation $w_i$:
\begin{equation}
	\mathcal L_{\mathrm{adapt}}
	=
	\frac{
		\sum_{i\in\mathcal M}
		w_i
		\left\|
		\mathcal D(z_i)-\bar x_i
		\right\|_2^2
	}{
		\sum_{i\in\mathcal M}w_i
	}.
	\label{eq:adaptation_loss}
\end{equation}
Here,
$\bar x_i=(x_i-\mu(x_i))/(\sigma(x_i)+\varepsilon)$ and
$w_i=\mathrm{clip}(\sigma(x_i),\sigma_{\min},\sigma_{\max})$.
Only the sonar pathway is optimized, and the decoder $\mathcal D$ is discarded afterward.

\paragraph{Stage II: Acoustic Semantic Learning.}
Category supervision then organizes the adapted sonar representations according to acoustic semantics:
\begin{equation}
	\begin{aligned}
		\mathcal L_{\mathrm{sem}}
		&=
		-\frac{1}{B}
		\sum_{n=1}^{B}
		\sum_{c=1}^{C}
		y_{n,c}\log p_{s,c}^{(n)},\\
		p_s^{(n)}
		&=
		\mathrm{softmax}\left(
		W_c h_s^{(n)}+b_c
		\right).
	\end{aligned}
	\label{eq:semantic_loss}
\end{equation}
where
$h_s^{(n)}=\mathrm{Pool}[\mathcal E_s(\widetilde I_s^{(n)})]$.
The optical pathway, language model, and fusion modules remain frozen.

\paragraph{Stage III: Cross-Modal Alignment and Reliability Learning.}
Synchronized sonar--optical pairs are used with stochastic optical degradation
across a continuous severity range, while clear pairs are retained as anchors.
Training combines global contrastive alignment,
hierarchical feature alignment, BCE-based gate supervision, and gate
regularization:
\begin{equation}
	\begin{aligned}
		\mathcal L_{\mathrm{NCE}}
		&=
		\mathrm{InfoNCE}(z_o,z_s),\\
		\mathcal L_{\mathrm{hier}}
		&=
		\sum_{k=1}^{3}
		\left[
		1-
		\cos\left(
		z_o^{(k)},z_s^{(k)}
		\right)
		\right],\\
		\mathcal L_{\mathrm{gate}}
		&=
		\frac{1}{4}
		\sum_{k=0}^{3}
		\mathrm{BCE}\left(
		g_s^{(k)},\pi_s(\eta)
		\right),\\
		\mathcal L_{\mathrm{ent}}
		&=
		-\frac{1}{4}
		\sum_{k=0}^{3}
		\mathcal H_b\!\left(g_s^{(k)}\right),\\
		\mathcal L_{\mathrm{align}}
		&=
		\mathcal L_{\mathrm{NCE}}
		+0.5\mathcal L_{\mathrm{hier}}
		+0.2\mathcal L_{\mathrm{gate}}
		+0.1\mathcal L_{\mathrm{ent}}.
	\end{aligned}
	\label{eq:alignment_loss}
\end{equation}
Here, $z_m$ and $z_m^{(k)}$ are pooled final and intermediate
representations, respectively; InfoNCE uses temperature $0.07$, and
$\mathcal H_b$ denotes binary entropy. The BCE target
$\pi_s(\eta)=\mathrm{clamp}(0.5+0.4\eta,0,1)$ shifts supervision from balanced
fusion toward sonar as optical degradation increases, making $g_s$ a supervised
degradation proxy rather than a general reliability estimate. The
negative-entropy term discourages early gate collapse.
Modality-missing examples preserve compatibility
with sonar-only and optical-only inputs.

\paragraph{Stage IV: Multimodal Instruction Tuning.}
Finally, the model is trained on sonar-only, optical-only, and paired sonar--optical instructions. For visual input
$\mathcal I\in\{I_s,I_o,(I_s,I_o)\}$,
prompt $x$, and target answer $y$, we minimize the answer-only autoregressive objective:
\begin{equation}
	\mathcal L_{\mathrm{inst}}
	=
	-\frac{1}{|\mathcal A|}
	\sum_{t\in\mathcal A}
	\log
	p_{\Theta}
	\left(
	y_t
	\mid
	x,\mathcal I,y_{<t}
	\right),
	\label{eq:instruction_loss}
\end{equation}
where $\mathcal A$ contains answer-token positions and $\Theta$ includes the trainable LoRA parameters~\citep{hu2021lora} and multimodal interfaces. This stage transfers the learned sonar representations and reliability-aware alignment to open-ended recognition, counting, question answering, and captioning.

\section{Experiments}
\label{sec:experiments}

Our evaluation follows the causal structure of SonarLLM. We first test whether
a sonar-native pathway provides capabilities that cannot be recovered through
model scale or instruction tuning alone. We then use paired optical degradation
to isolate when sonar becomes complementary and whether AGFM responds to the
resulting reliability shift. Finally, representation analysis and controlled
ablations trace these gains to sonar-domain adaptation, cross-modal alignment,
sonar geometry, modality-specific enhancement, and hierarchical interaction
before we quantify their computational cost. Unless otherwise specified, all
models use identical task prompts, input protocols, and deterministic decoding.

\subsection{Experimental Setup}
\label{sec:experimental_setup}

\paragraph{Training Data.}
Stage I pools approximately 98K candidate sonar images from RGBS50, UATD,
SCTD, DeeperSense, FLC+FLS, and OceanGym~\citep{xue2025oceangym}, together with
sonar-filtered frames from OceanInstruct and OceanPile's
OceanInstruction~\citep{xue2026oceanpile}. After empty-frame removal, outlier
filtering, byte-level deduplication, and brightness-balanced sampling, we
retain 40K unlabeled images for sonar-domain adaptation.
Stage II uses 23-class sonar object annotations aggregated from these sources.
Stage III uses synchronized RGBS50 sonar--optical pairs with online optical
degradation. Stage IV uses a 635K-sample instruction mixture comprising 533K
sonar-related and 102K optical samples, with sonar-only, optical-only, and
paired inputs.

\paragraph{SonarBench.}
SonarBench evaluates recognition, counting, VQA, and captioning. Recognition,
counting, and VQA use sonar-only, optical-only, and fusion inputs; captioning
uses sonar-only and optical-only inputs. Optical and fusion settings are
evaluated under clear, turbid, and heavily turbid conditions. Across
degradation levels, the scene, question, and sonar observation remain fixed,
and only the paired RGB image is modified. This design is a controlled stress
test of optical reliability rather than a simulation of the complete
distribution of natural turbidity. Within this protocol, paired fusion
analysis is defined over recognition, counting, and VQA, whereas captioning is
reported separately as a diagnostic of open-ended semantic generation.

\begin{table}[t]
	\centering
	\caption{SonarBench evaluation structure. S/O/F denote sonar, optical, and
		fusion inputs; C/T/H denote clear, turbid, and heavy optical conditions.}
	\label{tab:benchmark_protocol}
	\small
	\setlength{\tabcolsep}{2.8pt}
	\begin{tabular}{@{}lccc@{}}
		\toprule
		Task & Input conditions & Metric & Subsets \\
		\midrule
		Recognition & S; O/F$\times$C/T/H & Sem. accuracy & 7 \\
		Counting & S; O/F$\times$C/T/H & Exact accuracy & 7 \\
		VQA & S; O/F$\times$C/T/H & Sem. accuracy & 7 \\
		Captioning & S; O$\times$C/T/H & GOOD / METEOR & 4 \\
		\midrule
		Total & & & 25 \\
		\bottomrule
	\end{tabular}
	\vspace{-5pt}
\end{table}

The three optical conditions provide a controlled contrast in observation
quality. Clear images remain unchanged; turbid images combine reduced
brightness, additive white noise, colored veiling, and Gaussian blur; and
heavy-turbid images retain the same photometric attenuation while replacing
white noise with multi-scale correlated disturbances. Training degradation
data and benchmark renderings use the same generator, so the comparison
isolates response to a known reliability shift rather than out-of-family
generalization to arbitrary natural turbidity.

Evaluation samples are drawn from held-out RGBS50 and UMOD video sequences,
forming 25 subsets---4 sonar-only, 12 optical-only, and 9 fusion---with 150
QA instances each. Each record contains one question, while an underlying
sequence--frame may recur across tasks, modalities, and degradation conditions
to support paired comparisons. Source sequences are partitioned before all
training stages, and a sequence--frame audit confirms zero overlap between
training and evaluation.

\paragraph{Metrics and Judging.}
Recognition and VQA use Qwen3.8-Max~\citep{qwen2026qwen38} to judge semantic
correctness against reference answers. Counting uses integer parsing and exact
match; unparseable outputs (0.9\%) are incorrect. Captioning reports semantic
GOOD rate, with NLTK METEOR~\citep{banerjee2005meteor} ($\times100$) used only
as a lexical diagnostic. Macro averages cover recognition, counting, and VQA;
captioning is reported separately. VQA equally macro-averages attribute,
existence, and counting scores, so its values need not be integer multiples of
$1/150$; reported improvements are computed from unrounded aggregate scores.
To assess possible same-family judge bias,
we independently re-evaluate a stratified set of 300 recognition/VQA outputs
with Kimi-K3~\citep{moonshot2026kimik3}. The judges achieve 94.3\% agreement
and Cohen's $\kappa=0.87$, and all model rankings remain unchanged after manual
inspection of disagreements.

\paragraph{Baselines.}
We compare against Qwen3-VL-8B, InternVL3.5-8B~\citep{wang2025internvl35},
MiniCPM-V 4.5~\citep{yu2025minicpmv45}, the larger Qwen3.6-35B-A3B
and Qwen3.8-27B models~\citep{qwen2026qwen36,qwen2026qwen3827b}, and the
underwater-domain OceanGPT-o-7B and NAUTILUS-7B models. Qwen3-VL-8B+LoRA uses
the same Stage-IV data and LoRA configuration as SonarLLM, providing a
controlled test of whether instruction tuning alone explains the gains. For
fusion input, all baselines use their native multi-image interface with a fixed
sonar--optical order and explicit modality labels.

\paragraph{Implementation Details.}
SonarLLM uses Qwen3-VL-8B with $448\times448$ inputs. The language model is
frozen in Stages I--III. Stage I uses a 0.60 masking ratio. Stage II optimizes
PSVT and the Sonar Stem at a learning rate of $5\times10^{-5}$. Stage III
degrades the optical input with probability $0.7$, samples
$\eta\sim\mathcal U(0.1,1.0)$ for degraded examples, and jointly trains the
sonar pathway, both VFEs, and AGFM at a learning rate of $10^{-4}$. Stage IV uses LoRA
with $r=128$, $\alpha=256$, and learning rate $10^{-4}$. The architecture and
Stage-IV adapters add 882.4M and 349.2M parameters, respectively, resulting in
10.3B total parameters versus 8.77B for the backbone when the frozen
DINOv2-L structural-prior encoder is included.

\subsection{Main Results}
\label{sec:main_results}

\begin{table*}[t]
	\centering
	\caption{
		SonarBench results (\%; 25 subsets, $n{=}150$ each).
		Recognition/VQA use semantic judging, counting uses exact match, and
		$^\ddagger$ denotes caption GOOD rate. Averages exclude captioning. Best per
		row in bold.
	}
	\label{tab:main}
	\small
	\setlength{\tabcolsep}{3pt}
	\begin{tabular}{ll ccccccccc}
		\toprule
		Modality & Task (Condition)
		& \shortstack{Qwen3-VL\\-8B}
		& \shortstack{Qwen3-VL\\-8B+LoRA}
		& \shortstack{InternVL\\3.5-8B}
		& \shortstack{MiniCPM-V\\4.5}
		& \shortstack{NAUTILUS\\-7B}
		& \shortstack{OceanGPT\\-o-7B}
		& \shortstack{Qwen3.6\\-35B-A3B}
		& \shortstack{Qwen3.8\\-27B}
		& \shortstack{\textbf{Sonar}\\\textbf{LLM}} \\
		\midrule
		
		\raisebox{-1.5\baselineskip}[0pt][0pt]{\textit{Sonar}}
		& Recognition
		& 25.3 & 28.7 & 29.3 & 23.3 & 14.0 & 2.0
		& 27.3 & 32.0 & \textbf{64.7} \\
		
		&
		Counting
		& 40.7 & 51.3 & 33.3 & 47.3 & 32.7 & 14.0
		& 35.3 & 32.7 & \textbf{84.7} \\
		
		&
		VQA
		& 30.2 & 32.6 & 27.5 & 21.7 & 25.8 & 20.8
		& 24.2 & 22.5 & \textbf{66.5} \\
		
		&
		Caption$^\ddagger$
		& 8.0 & 4.0 & 7.3 & 4.7 & 12.0 & 5.3
		& 10.0 & 4.7 & \textbf{33.3} \\
		
		\midrule
		
		\raisebox{-5.5\baselineskip}[0pt][0pt]{\textit{Optical}}
		& Recognition (Clear)
		& 41.3 & 38.7 & 38.7 & 36.0 & 34.7 & 10.7
		& 44.0 & 40.0 & \textbf{77.3} \\
		
		&
		Recognition (Turbid)
		& 37.3 & 36.0 & 36.0 & 35.3 & 36.0 & 12.7
		& 40.7 & 38.7 & \textbf{63.3} \\
		
		&
		Recognition (Heavy)
		& 38.0 & 35.3 & 33.3 & 29.3 & 30.7 & 9.3
		& 34.7 & 39.3 & \textbf{44.7} \\
		
		&
		Counting (Clear)
		& 71.3 & 69.3 & 60.7 & 66.0 & 59.3 & 40.7
		& \textbf{73.3} & 72.7 & 66.0 \\
		
		&
		Counting (Turbid)
		& 37.3 & \textbf{39.3} & 26.0 & 32.0 & 24.0 & 22.7
		& 29.3 & 22.7 & 37.3 \\
		
		&
		Counting (Heavy)
		& 8.0 & 16.0 & 6.7 & 10.0 & 6.0 & 10.0
		& 10.0 & 6.0 & \textbf{36.0} \\
		
		&
		VQA (Clear)
		& 40.9 & 37.0 & 37.0 & 41.1 & 32.6 & 22.9
		& 38.7 & 36.8 & \textbf{47.5} \\
		
		&
		VQA (Turbid)
		& 34.7 & 32.5 & 30.8 & 32.0 & 28.2 & 21.5
		& 32.8 & 28.5 & \textbf{44.7} \\
		
		&
		VQA (Heavy)
		& 28.1 & 28.2 & 26.3 & 22.9 & 20.7 & 20.3
		& 24.1 & 24.7 & \textbf{41.2} \\
		
		&
		Caption (Clear)$^\ddagger$
		& 23.3 & 26.7 & 25.3 & 22.0 & 18.7 & 16.0
		& 26.7 & 22.0 & \textbf{29.3} \\
		
		&
		Caption (Turbid)$^\ddagger$
		& 24.7 & 26.0 & 24.7 & 25.3 & 22.7 & 12.0
		& 24.7 & 21.3 & \textbf{28.7} \\
		
		&
		Caption (Heavy)$^\ddagger$
		& \textbf{15.3} & 9.3 & 10.0 & 10.7 & 14.0 & 6.0
		& 13.3 & 12.0 & 14.7 \\
		
		\midrule
		
		\raisebox{-4\baselineskip}[0pt][0pt]{\textit{Fusion}}
		& Recognition (Clear)
		& 37.3 & 30.7 & 34.7 & 34.7 & 34.0 & 15.3
		& 40.7 & 40.7 & \textbf{83.3} \\
		
		&
		Recognition (Turbid)
		& 36.0 & 30.0 & 36.0 & 39.3 & 33.3 & 15.3
		& 38.7 & 40.0 & \textbf{76.7} \\
		
		&
		Recognition (Heavy)
		& 39.3 & 30.0 & 33.3 & 32.0 & 30.7 & 10.7
		& 37.3 & 38.7 & \textbf{80.7} \\
		
		&
		Counting (Clear)
		& 66.7 & 36.0 & 65.3 & 62.0 & 62.7 & 52.7
		& 23.3 & 63.3 & \textbf{72.0} \\
		
		&
		Counting (Turbid)
		& 44.7 & 36.0 & 59.3 & 51.3 & 63.3 & 52.0
		& 25.3 & 61.3 & \textbf{70.7} \\
		
		&
		Counting (Heavy)
		& 47.3 & 36.0 & 48.7 & 36.0 & 62.0 & 55.3
		& 24.0 & 40.0 & \textbf{72.0} \\
		
		&
		VQA (Clear)
		& 39.1 & 30.3 & 33.9 & 38.0 & 38.0 & 11.6
		& 25.6 & 38.5 & \textbf{54.6} \\
		
		&
		VQA (Turbid)
		& 33.7 & 31.5 & 34.4 & 26.9 & 28.9 & 15.5
		& 24.7 & 36.1 & \textbf{53.5} \\
		
		&
		VQA (Heavy)
		& 26.4 & 30.1 & 28.6 & 20.4 & 22.6 & 13.5
		& 20.3 & 34.5 & \textbf{55.2} \\
		
		\midrule
		
		\multicolumn{2}{l}{Sonar Average}
		& 32.1 & 37.5 & 30.0 & 30.8 & 24.2 & 12.3
		& 28.9 & 29.1 & \textbf{72.0} \\
		
		\multicolumn{2}{l}{Optical Average}
		& 37.4 & 36.9 & 32.8 & 33.8 & 30.2 & 19.0
		& 36.4 & 34.4 & \textbf{50.9} \\
		
		\multicolumn{2}{l}{Fusion Average}
		& 41.2 & 32.3 & 41.6 & 37.8 & 41.7 & 26.9
		& 28.9 & 43.7 & \textbf{68.7} \\
		
		\bottomrule
	\end{tabular}
\end{table*}

\paragraph{Native Sonar Understanding.}
SonarLLM obtains 72.0\% sonar-only macro accuracy, exceeding the strongest
baseline by 34.4 points. The gain is consistent across recognition (64.7\%),
counting (84.7\%), and VQA (66.5\%), and its caption GOOD rate reaches 33.3\%
versus 12.0\% for the best baseline. Increasing general-purpose capacity does
not close the gap: Qwen3.6-35B-A3B and Qwen3.8-27B achieve only 28.9\% and
29.1\%, respectively. Underwater-domain models are also substantially weaker,
supporting the need for sonar-native representation rather than model scaling
or domain knowledge alone.

\paragraph{Optical and Fusion Performance.}
SonarLLM achieves 50.9\% on optical input and ranks first on every recognition
and VQA subset. Its main optical weakness is clear/turbid counting, where it
trails the best baselines. With both sensors, however, SonarLLM reaches 68.7\%,
25.1 points above the strongest baseline. Fusion recognition remains between
76.7\% and 83.3\% across all degradation levels. In contrast, heterogeneous
multi-image input alone is unreliable: Qwen3-VL-8B+LoRA drops from 36.9\%
optical accuracy to 32.3\% under fusion, and Qwen3.6-35B-A3B drops from 36.4\%
to 28.9\%. Dedicated heterogeneous representation and interaction therefore
provide substantial benefits when exploiting the second sensor.

\paragraph{Controlling Instruction Tuning and Sampling Uncertainty.}
Qwen3-VL-8B+LoRA uses the same Stage-IV data and language-side LoRA as
SonarLLM. Relative to this control, SonarLLM improves sonar, optical, and fusion
accuracy by 34.4, 14.0, and 36.5 points, showing that instruction tuning alone
does not reproduce the observed gains. A 10,000-replication scene-clustered
bootstrap gives 95\%
confidence intervals of $[28.1,40.5]$, $[11.3,16.7]$, and $[34.0,39.0]$ for
these gains; all lower bounds remain positive. These intervals quantify
test-set sampling uncertainty but not variation across independent training
runs.

\subsection{Controlled Degradation and Reliability-Aware Fusion}
\label{sec:robustness}
\label{sec:gate_analysis}

Because optical and fusion evaluations share scenes, questions, and sonar
observations, fusion minus optical accuracy provides a paired estimate of the
benefit of adding sonar. Table~\ref{tab:degradation_gain} summarizes this controlled
comparison. The fusion-over-optical gain expands from 6.0 to 36.0 points for
recognition and counting, and from 7.1 to 14.0 points for VQA. Meanwhile,
clear-to-heavy fusion changes are limited to $-2.6$, $0.0$, and $+0.6$ points,
respectively, despite much larger optical declines. For recognition, fusion
also exceeds the stronger individual sensor by 6.0, 12.0, and 16.0 points
across clear, turbid, and heavy conditions, showing that sonar increasingly
complements rather than merely replaces optical sensing.

\begin{table}[t]
	\centering
	\caption{Controlled degradation summary derived from Table~\ref{tab:main}
		(percentage points). $\Delta$(F--O) is fusion minus optical accuracy;
		C/T/H denote clear/turbid/heavy conditions.}
	\vspace{-5pt}
	\label{tab:degradation_gain}
	\small
	\setlength{\tabcolsep}{2.8pt}
	\begin{tabular}{lccc}
		\toprule
		Task & \shortstack{$\Delta$(F--O)\\C/T/H}
		& \shortstack{Optical\\C$\rightarrow$H}
		& \shortstack{Fusion\\C$\rightarrow$H} \\
		\midrule
		Recognition & $+6.0/+13.4/+36.0$ & $-32.6$ & $-2.6$ \\
		Counting & $+6.0/+33.4/+36.0$ & $-30.0$ & $0.0$ \\
		VQA & $+7.1/+8.8/+14.0$ & $-6.3$ & $+0.6$ \\
		\bottomrule
	\end{tabular}
	\vspace{-5pt}
\end{table}

\begin{figure*}[t]
	\centering
	\includegraphics[width=0.82\textwidth]{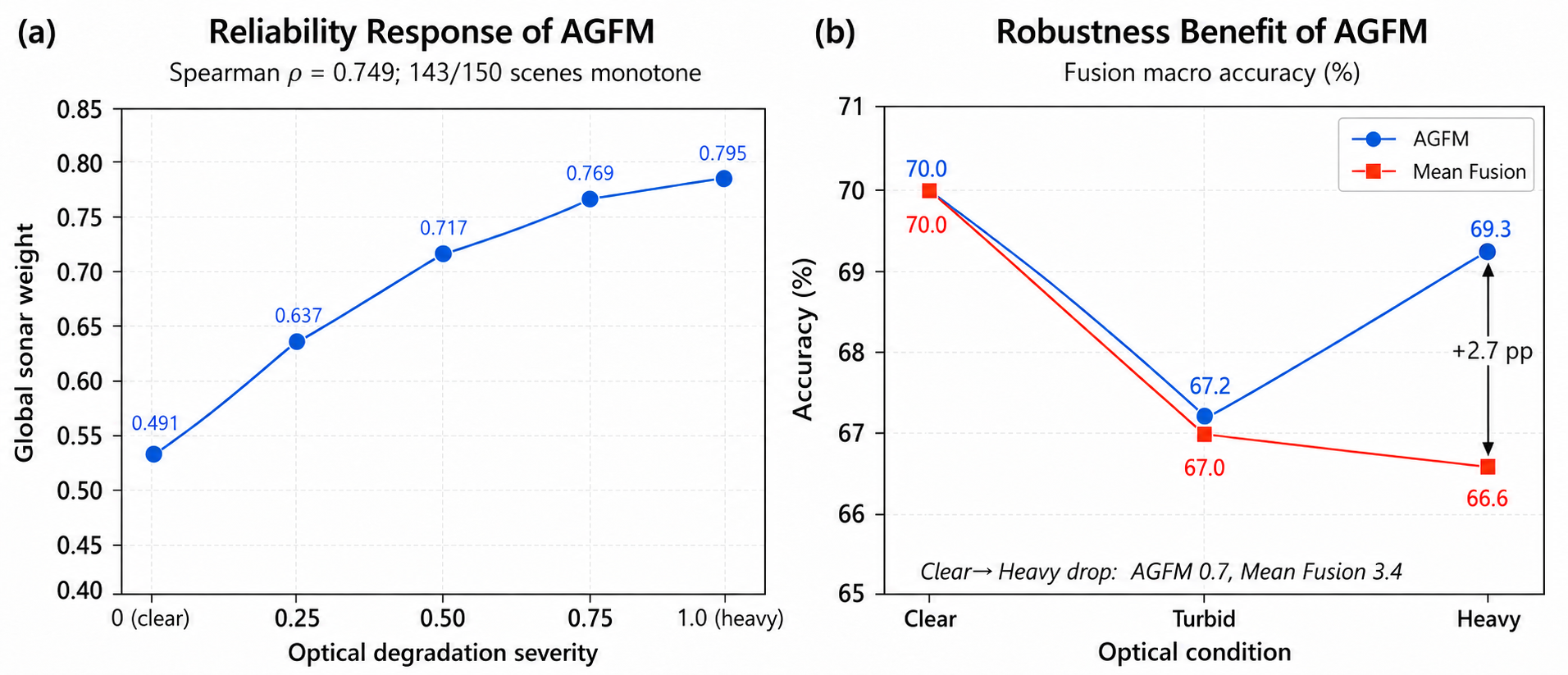}
	\caption{Reliability-aware fusion analysis on 150 held-out scenes.
		\textbf{(a)} Sonar weight increases with optical degradation.
		\textbf{(b)} Under the same checkpoint, AGFM's advantage over equal weighting
		grows from 0.0 to 2.7 points.}
	\label{fig:agfm_analysis}
	\vspace{-10pt}
\end{figure*}

We next evaluate whether AGFM exhibits the intended internal response on an
independent probe set of 150 RGBS50 scenes, each rendered at five degradation
levels ($750$ observations). As shown in Fig.~\ref{fig:agfm_analysis}(a), mean
sonar weight increases from 0.491 to 0.637, 0.717, 0.769, and 0.795. Severity
and sonar weight have Spearman correlation $\rho=0.749$ (scene-blocked 95\% CI
$[0.68,0.81]$), and 95.3\% of scenes show non-decreasing sonar weight.
Because Stage III explicitly supervises the gate using degradation strength,
we interpret this trend as verification of AGFM's intended internal response,
rather than as independent evidence of emergent reliability estimation.

Using the same checkpoint, we replace AGFM by equal weighting at inference.
The accuracy advantage is 0.0, 0.2, and 2.7 points under clear, turbid, and
heavy conditions, respectively; clear-to-heavy degradation is 0.7 points with
AGFM versus 3.4 points with equal weighting. This pattern clarifies that
complementarity is not an unconditional advantage of using more sensors.
Recognition benefits consistently because optical appearance and acoustic
contour and range provide distinct evidence, whereas fusion counting and VQA
do not always exceed sonar-only performance when the answer depends primarily
on geometry or degraded optical cues become distracting. AGFM is therefore not
an oracle that selects the best modality for every question; it acts mainly as
a degradation buffer that prevents a deteriorating sensor from dominating the
shared representation. This distinction separates representation from
allocation: the sonar pathway determines which acoustic evidence is available
to language reasoning, while AGFM controls its influence as observation
reliability shifts. The following ablations examine these responsibilities at
the representation and interaction levels.
The controlled protocol does not cover non-uniform scattering, dynamic
suspended particles, sonar-specific failures, or temporal misalignment;
evaluation on naturally degraded paired observations remains necessary.

\subsection{Ablation and Representation Analysis}
\label{sec:ablation}

Ablations use a diagnostic split constructed from video sequences disjoint
from both training and the main SonarBench test set, with 25
modality--task--condition combinations and 150 samples each ($N=3{,}750$).
Recognition and VQA use rule-based scoring rather than the semantic judge in
Table~\ref{tab:main}; absolute values across the two tables are not directly
comparable.

\paragraph{Domain Adaptation and Representation Formation.}
\begin{table}[t]
	\centering
	\caption{Sonar front-end adaptation on the rule-scored validation split.
		R/C avg. is the mean of recognition and counting accuracy; METEOR is
		reported in \%.}
	\vspace{-5pt}
	\label{tab:frontend_ablation}
	\small
	\setlength{\tabcolsep}{2.8pt}
	\begin{tabular}{@{}lcccc@{}}
		\toprule
		Variant & R/C avg. & Rec. & Count & METEOR \\
		\midrule
		Transplant (no MAE) & 72.0 & 64.0 & \textbf{80.0} & 45.1 \\
		MAE-$r64$ & 69.3 & 61.3 & 77.3 & 46.8 \\
		\textbf{MAE-$r128$} & \textbf{74.7} & \textbf{71.3} & 78.0 & \textbf{49.2} \\
		\bottomrule
	\end{tabular}
	\vspace{-5pt}
\end{table}

As shown in Table~\ref{tab:frontend_ablation}, Stage-I MAE with the same
$r=128$ language adaptation improves the sonar R/C average from 72.0\% to 74.7\%,
recognition by 7.3 points, and METEOR by 4.1, although counting decreases by
2.0 points. MAE-$r64$ reaches only 69.3\%, indicating that adapted acoustic
features also require sufficient language-side capacity. Mask ratios
and Stage-I data scale are examined separately in
Table~\ref{tab:stage1_sensitivity}.

\begin{table}[t]
	\centering
	\caption{Sensitivity of the sonar R/C average (\%) to Stage-I masking and data scale.}
	\vspace{-5pt}
	\label{tab:stage1_sensitivity}
	\small
	\setlength{\tabcolsep}{4.0pt}
	\begin{tabular}{lcc}
		\toprule
		Factor & Tested values & R/C avg. \\
		\midrule
		Mask ratio & 0.50 / 0.60 / 0.75 & 73.3 / \textbf{74.7} / 69.3 \\
		Unlabeled images & 10K / 20K / 40K & 70.0 / 72.7 / \textbf{74.7} \\
		\bottomrule
	\end{tabular}
	\vspace{-10pt}
\end{table}

An intermediate masking ratio performs best: excessive masking removes sparse
acoustic structure, whereas insufficient masking weakens the adaptation
signal. Increasing unlabeled data produces consistent but diminishing gains;
we therefore use a 0.60 mask ratio and 40K images.

\begin{figure*}[t]
	\centering
	\includegraphics[width=0.98\textwidth]{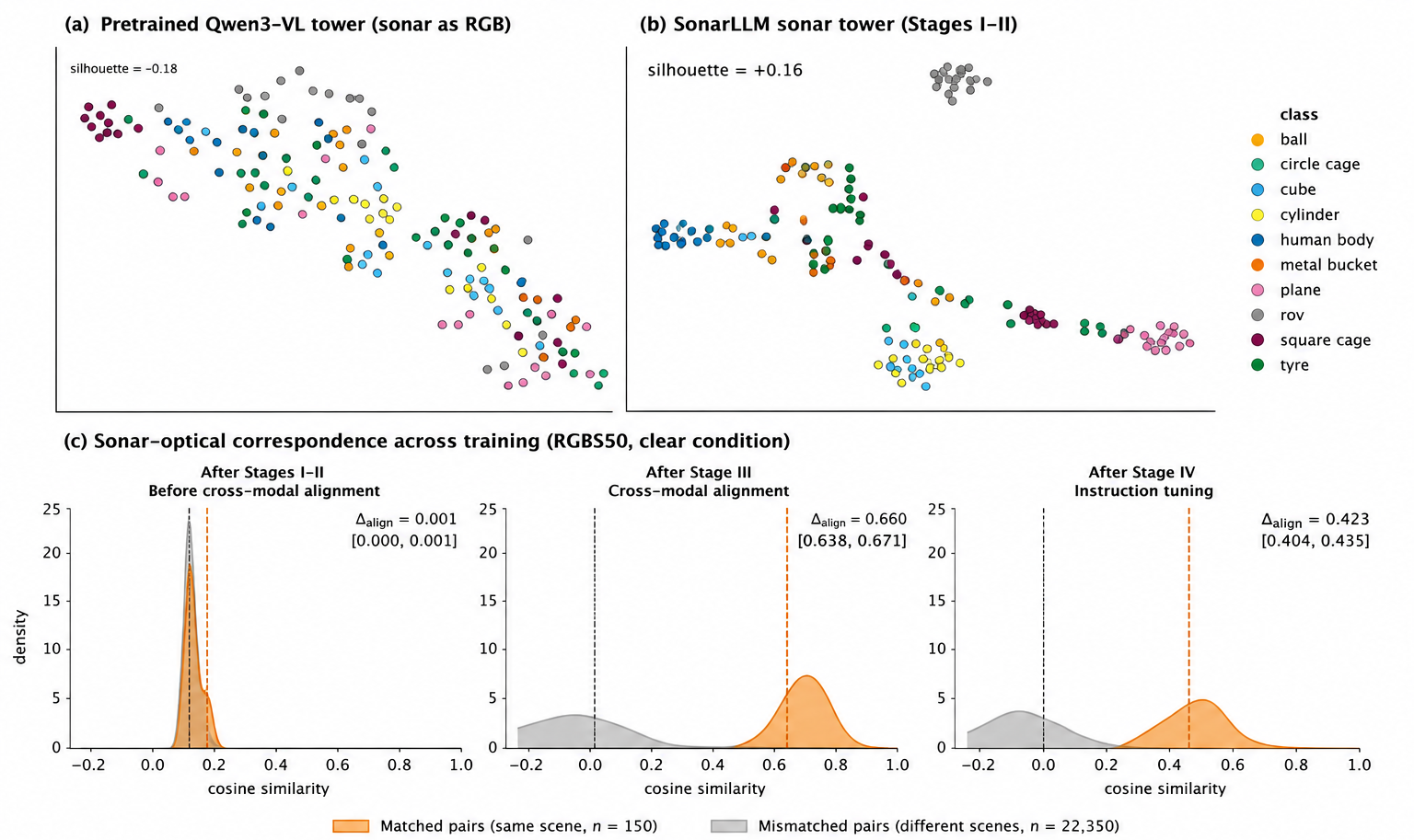}
	\caption{Progressive representation formation. \textbf{(a)} Sonar frames
		processed by the pretrained optical tower. \textbf{(b)} Class structure after
		Stages I--II. \textbf{(c)} Similarity of matched and mismatched sonar--optical
		pairs before and after alignment and instruction tuning. Panel (c) uses 150
		matched same-scene pairs and 22,350 mismatched cross-scene pairs under the
		clear condition; intervals
		denote 95\% confidence intervals for the mean-similarity margin.}
	\label{fig:representation_alignment}
\end{figure*}

Figure~\ref{fig:representation_alignment} shows that the silhouette score of
sonar categories increases from $-0.18$ under the pretrained optical tower to
$0.16$ after Stages I--II. However, matched and mismatched sonar--optical pairs
remain nearly indistinguishable at this point
($\Delta_{\mathrm{align}}=0.001$, 95\% CI $[0.000,0.001]$). Stage III raises
the alignment margin to 0.660 ($[0.638,0.671]$), supporting the conclusion that
cross-modal correspondence is learned beyond unimodal sonar structure. After
Stage IV, the margin remains substantial at 0.423 ($[0.404,0.435]$), indicating
that instruction tuning relaxes strict feature similarity while preserving
task-relevant correspondence.

\paragraph{Structural Components.}
\begin{table}[t]
	\centering
	\caption{Structural ablation on the rule-scored validation split (\%).
		Fusion, Optical, and Sonar Macro average recognition, counting, and VQA.
		\emph{Drop}: clear-to-heavy fusion decline. $\dagger$: sonar front-end
		retraining required. Mean Fusion is an independently retrained
		equal-weight variant.}
	\vspace{-8pt}
	\label{tab:module_ablation}
	\small
	\setlength{\tabcolsep}{3.5pt}
	\begin{tabular}{lcccc}
		\toprule
		Variant & Fusion & Drop & Optical & \shortstack{Sonar\\Macro} \\
		\midrule
		\textbf{SonarLLM}
		& \textbf{82.1} & \textbf{1.3}
		& \textbf{72.5} & \textbf{65.8} \\
		Mean Fusion
		& 80.8 & 3.8 & 72.3 & 65.6 \\
		w/o gate loss
		& 81.2 & 3.0 & 72.7 & 64.9 \\
		w/o both VFE
		& 79.8 & 3.1 & 67.2 & 61.4 \\
		w/o Optical-VFE
		& 80.7 & 2.6 & 68.8 & 65.4 \\
		w/o Acoustic-VFE
		& 80.4 & 2.4 & 72.1 & 62.0 \\
		w/o DeepStack
		& 78.9 & 2.8 & 71.8 & 59.6 \\
		w/o Sonar Stem$^\dagger$
		& 73.9 & 5.3 & 71.6 & 58.3 \\
		w/o Polar PE$^\dagger$
		& 68.6 & 7.1 & 71.4 & 55.7 \\
		\bottomrule
	\end{tabular}
	\vspace{-15pt}
\end{table}

Table~\ref{tab:module_ablation} identifies sonar geometry as the dominant
factor: removing Polar PE reduces fusion/sonar-macro accuracy by 13.5/10.1 points,
and removing the Sonar Stem reduces them by 8.2/7.5 points. The VFEs exhibit
the intended modality specificity: removing Optical-VFE mainly reduces
optical accuracy (3.7 points), whereas removing Acoustic-VFE mainly reduces
sonar-macro accuracy (3.8 points). Removing DeepStack causes a 6.2-point sonar-macro loss,
showing the importance of intermediate acoustic structure. Removing both VFEs
causes consistent losses of 2.3, 5.3, and 4.4 points on fusion, optical, and
sonar inputs, respectively, confirming complementary rather than redundant
modality-specific corrections.

Quality-aware weighting contributes most strongly to robustness. Mean Fusion and
w/o gate loss reduce static fusion accuracy by only 1.3 and 0.9 points, but
increase the clear-to-heavy decline from 1.3 to 3.8 and 3.0 points. Together
with the same-checkpoint comparison in Fig.~\ref{fig:agfm_analysis}, this
consistently supports AGFM as a quality-aware adaptation mechanism. The Sonar
Stem and Polar PE variants require front-end retraining and are therefore not
strict inference-time ablations, but their consistent losses across sonar,
fusion, and degradation sensitivity support their necessity.

Together, the ablations reveal complementary roles: the Sonar Stem and Polar PE
establish acoustic representations, the VFEs correct modality-specific
corruption, DeepStack preserves intermediate structure, and AGFM stabilizes
fusion under asymmetric observation quality.

\subsection{Efficiency}
\label{sec:efficiency}

\begin{table}[t]
	\vspace{-10pt}
	\centering
	\caption{Inference efficiency on one A100-80GB GPU. Results use BF16,
		FlashAttention-2, batch size 1, two $448\times448$ images, 512 visual tokens, and 128
		generated tokens; all SonarLLM measurements  and latency and throughput are mean$\pm$std over 10 runs.}
	\label{tab:efficiency}
	\small
	\setlength{\tabcolsep}{2.3pt}
	\begin{tabular}{lcccc}
		\toprule
		Model & Params & \shortstack{Peak\\mem.}
		& \shortstack{Prefill\\(ms)} & \shortstack{Decode\\(tok/s)} \\
		\midrule
		Qwen3-VL-8B & 8.77B & 17.7GB & $77.2{\pm}0.2$ & $38.4{\pm}0.1$ \\
		SonarLLM & 10.3B & 21.6GB & $137.5{\pm}1.1$ & $21.1{\pm}1.2$ \\
		\bottomrule
	\end{tabular}
	\vspace{-15pt}
\end{table}

Table~\ref{tab:efficiency} shows the cost of native sonar processing:
parameters and peak memory rise by 17.4\% and 22.0\%, prefill latency by 78.1\%,
and decoding throughput falls by 45.1\%. Most added capacity is inherited: the
736.9M sonar tower is pretrained and frozen DINOv2-L contributes 0.30B
parameters; only 25.1M non-LoRA adaptation and fusion parameters are randomly
initialized.

\section{Conclusion}
\label{sec:conclusion}

This work shows that robust sonar--optical reasoning requires representations
tailored to each modality and fusion conditioned on observation quality.
SonarLLM combines a sonar-native pathway with SonarBench's paired protocol,
which varies optical quality while holding the scene and sonar observation
fixed. It achieves 72.0\% sonar-only macro accuracy and 68.7\% fusion accuracy,
34.4 and 25.1 points above the strongest baselines. The fusion-over-optical gain
rises from $+6.0$ to $+36.0$ points as optical quality deteriorates, showing
that sonar becomes increasingly valuable as optical reliability falls.

These results support representing heterogeneous sensors according to their
observation processes and weighting them according to observation quality. The
controlled protocol does not fully validate naturally occurring turbidity;
future work will extend to real paired data, task-aware routing, and temporal
sonar--optical reasoning.

\bibliographystyle{ACM-Reference-Format}
\bibliography{aaai2027}

@misc{zheng2023marinegpt,
  author        = {Zheng, Ziqiang and Zhang, Jipeng and Vu, Tuan-Anh and Diao, Shizhe and {Yue Him Wong Tim} and Yeung, Sai-Kit},
  title         = {MarineGPT: Unlocking Secrets of Ocean to the Public},
  year          = {2023},
  eprint        = {2310.13596},
  archivePrefix = {arXiv}
}

@inproceedings{xu2025nautilus,
  author    = {Xu, Wei and Wang, Cheng and Liang, Dingkang and Zhao, Zongchuang and Jiang, Xingyu and Zhang, Peng and Bai, Xiang},
  title     = {NAUTILUS: A Large Multimodal Model for Underwater Scene Understanding},
  booktitle = {Advances in Neural Information Processing Systems (NeurIPS)},
  year      = {2025}
}

@inproceedings{bi2024oceangpt,
  author    = {Bi, Zhen and Zhang, Ningyu and Xue, Yida and Ou, Yixin and Ji, Daxiong and Zheng, Guozhou and Chen, Huajun},
  title     = {OceanGPT: A Large Language Model for Ocean Science Tasks},
  booktitle = {Proceedings of the 62nd Annual Meeting of the Association for Computational Linguistics (Volume 1: Long Papers)},
  pages     = {3357--3372},
  year      = {2024},
  publisher = {Association for Computational Linguistics},
  doi       = {10.18653/v1/2024.acl-long.184},
  url       = {https://aclanthology.org/2024.acl-long.184/}
}

@misc{alawode2025aquaticclip,
  author        = {Alawode, Basit Olakunle and Ganapathi, Iyyakutti Iyappan and Javed, Sajid and Werghi, Naoufel and Bennamoun, Mohammed and Mahmood, Arif},
  title         = {AquaticCLIP: A Vision-Language Foundation Model for Underwater Scene Analysis},
  year          = {2025},
  eprint        = {2502.01785},
  archivePrefix = {arXiv}
}

@article{neupane2020sonarreview,
  author    = {Neupane, Dhiraj and Seok, Jongwon},
  title     = {A Review on Deep Learning-Based Approaches for Automatic Sonar Target Recognition},
  volume    = {9},
  number    = {11},
  pages     = {1972},
  year      = {2020},
  publisher = {MDPI},
  journal   = {Electronics},
  doi       = {10.3390/electronics9111972},
  url       = {https://www.mdpi.com/2079-9292/9/11/1972}
}

@article{steiniger2022sonarsurvey,
  author    = {Steiniger, Yannik and Kraus, Dieter and Meisen, Tobias},
  title     = {Survey on Deep Learning Based Computer Vision for Sonar Imagery},
  volume    = {114},
  pages     = {105157},
  year      = {2022},
  publisher = {Elsevier},
  journal   = {Engineering Applications of Artificial Intelligence}
}

@article{xie2022uatd,
  author    = {Xie, Kaibing and Yang, Jian and Qiu, Kang},
  title     = {A Dataset with Multibeam Forward-Looking Sonar for Underwater Object Detection},
  volume    = {9},
  number    = {1},
  pages     = {739},
  year      = {2022},
  publisher = {Nature Publishing Group},
  journal   = {Scientific Data},
  doi       = {10.1038/s41597-022-01854-w},
  url       = {https://www.nature.com/articles/s41597-022-01854-w}
}

@article{zhang2022sctd,
  author    = {Zhang, Peng and Tang, Jinsong and Zhong, Heping and Ning, Mingqiang and Liu, Dandan and Wu, Ke},
  title     = {Self-Trained Target Detection of Radar and Sonar Images Using Automatic Deep Learning},
  volume    = {60},
  pages     = {4701914},
  year      = {2022},
  publisher = {IEEE},
  journal   = {IEEE Transactions on Geoscience and Remote Sensing},
  doi       = {10.1109/TGRS.2021.3096011}
}

@article{li2025rgbs50,
  author    = {Li, Yunfeng and Wang, Bo and Sun, Jiuran and Wu, Xueyi and Li, Ye},
  title     = {RGB-Sonar Tracking Benchmark and Spatial Cross-Attention Transformer Tracker},
  volume    = {35},
  number    = {3},
  pages     = {2260--2275},
  year      = {2025},
  publisher = {IEEE},
  journal   = {IEEE Transactions on Circuits and Systems for Video Technology},
  doi       = {10.1109/TCSVT.2024.3497214}
}

@article{wu2026umod,
  author    = {Wu, Yujie and Wang, Wenling and Lin, Cong and Hou, Mingxin and Liu, Mingxin},
  title     = {Towards Multimodal Underwater Object Detection: A Bidirectional Feature Recomposition Network and Visual-Sonar Dataset},
  volume    = {316},
  pages     = {131710},
  year      = {2026},
  publisher = {Elsevier},
  journal   = {Expert Systems with Applications},
  doi       = {10.1016/j.eswa.2026.131710}
}

@misc{chen2026sovis,
  author        = {Chen, Weitung and Tinn, Phil and Auran, Per Gunnar and Ludvigsen, Martin and Haro, Peter Halland},
  title         = {A Sonar-Visual Dataset for Cross-Modal Underwater Robot Perception},
  year          = {2026},
  eprint        = {2606.01398},
  archivePrefix = {arXiv},
  primaryClass  = {cs.RO},
  doi           = {10.48550/arXiv.2606.01398}
}

@misc{bai2025qwen3vl,
  author        = {Bai, Shuai and Cai, Yuxuan and Chen, Ruizhe and Chen, Keqin and Chen, Xionghui and Cheng, Zesen and Deng, Lianghao and Ding, Wei and Gao, Chang and Ge, Chunjiang and others},
  title         = {Qwen3-VL Technical Report},
  year          = {2025},
  eprint        = {2511.21631},
  archivePrefix = {arXiv}
}

@inproceedings{liu2023visualinstruction,
  author    = {Liu, Haotian and Li, Chunyuan and Wu, Qingyang and Lee, Yong Jae},
  title     = {Visual Instruction Tuning},
  booktitle = {Advances in Neural Information Processing Systems (NeurIPS)},
  volume    = {36},
  year      = {2023}
}

@misc{liu2023llava15,
  author        = {Liu, Haotian and Li, Chunyuan and Li, Yuheng and Lee, Yong Jae},
  title         = {Improved Baselines with Visual Instruction Tuning},
  year          = {2023},
  eprint        = {2310.03744},
  archivePrefix = {arXiv}
}

@inproceedings{li2023blip2,
  author    = {Li, Junnan and Li, Dongxu and Savarese, Silvio and Hoi, Steven},
  title     = {BLIP-2: Bootstrapping Language-Image Pre-training with Frozen Image Encoders and Large Language Models},
  booktitle = {Proceedings of the 40th International Conference on Machine Learning (ICML)},
  pages     = {19730--19742},
  year      = {2023},
  publisher = {PMLR}
}

@inproceedings{alayrac2022flamingo,
  author    = {Alayrac, Jean-Baptiste and Donahue, Jeff and Luc, Pauline and Miech, Antoine and Barr, Iain and Hasson, Yana and Lenc, Karel and Mensch, Arthur and Millican, Katie and Reynolds, Malcolm and others},
  title     = {Flamingo: A Visual Language Model for Few-Shot Learning},
  booktitle = {Advances in Neural Information Processing Systems (NeurIPS)},
  volume    = {35},
  pages     = {23716--23736},
  year      = {2022}
}

@inproceedings{arevalo2017gated,
  author    = {Arevalo, John and Solorio, Thamar and Montes-y-Gómez, Manuel and González, Fabio A.},
  title     = {Gated Multimodal Units for Information Fusion},
  booktitle = {ICLR 2017 Workshop Track},
  year      = {2017}
}

@article{liu2020uieb,
  author    = {Li, Chongyi and Guo, Chunle and Ren, Wenqi and Cong, Runmin and Hou, Junhui and Kwong, Sam and Tao, Dacheng},
  title     = {An Underwater Image Enhancement Benchmark Dataset and Beyond},
  volume    = {29},
  pages     = {4376--4389},
  year      = {2020},
  publisher = {IEEE},
  journal   = {IEEE Transactions on Image Processing},
  doi       = {10.1109/TIP.2019.2955241}
}

@article{islam2020euvp,
  author    = {Islam, Muhammad Jamil and Xia, Youya and Sattar, Junaed},
  title     = {Fast Underwater Image Enhancement for Improved Visual Perception},
  volume    = {5},
  number    = {2},
  pages     = {3227--3234},
  year      = {2020},
  publisher = {IEEE},
  journal   = {IEEE Robotics and Automation Letters}
}

@misc{xue2026oceanpile,
  author        = {Xue, Yida and Zhang, Ningyu and Wu, Tingwei and Ma, Zhe and Ji, Daxiong and Wang, Zhao and Zheng, Guozhou and Chen, Huajun},
  title         = {OceanPile: A Large-Scale Multimodal Ocean Corpus for Foundation Models},
  year          = {2026},
  eprint        = {2605.00877},
  archivePrefix = {arXiv}
}

@misc{xue2025oceangym,
  author        = {Xue, Yida and Mao, Mingjun and Ru, Xiangyuan and Zhu, Yuqi and Ren, Baochang and Qiao, Shuofei and Wang, Mengru and Deng, Shumin and An, Xinyu and Zhang, Ningyu and Chen, Ying and Chen, Huajun},
  title         = {OceanGym: A Benchmark Environment for Underwater Embodied Agents},
  year          = {2025},
  eprint        = {2509.26536},
  archivePrefix = {arXiv}
}

@misc{zhang2025uwbench,
  author        = {Zhang, Da and Rong, Chenggang and Li, Bingyu and Wang, Feiyu and Zhao, Zhiyuan and Gao, Junyu and Li, Xuelong},
  title         = {UWBench: A Comprehensive Vision-Language Benchmark for Underwater Understanding},
  year          = {2025},
  eprint        = {2510.18262},
  archivePrefix = {arXiv}
}

@misc{oquab2023dinov2,
  author        = {Oquab, Maxime and Darcet, Timoth{\'e}e and Moutakanni, Th{\'e}o and Vo, Huy and Szafraniec, Marc and Khalidov, Vasil and Fernandez, Pierre and Haziza, Daniel and Massa, Francisco and El-Nouby, Alaaeldin and Assran, Mahmoud and Ballas, Nicolas and Galuba, Wojciech and Howes, Russell and Huang, Po-Yao and Li, Shang-Wen and Misra, Ishan and Rabbat, Michael and Sharma, Vasu and Synnaeve, Gabriel and Xu, Hu and J{\'e}gou, Herv{\'e} and Mairal, Julien and Labatut, Patrick and Joulin, Armand and Bojanowski, Piotr},
  title         = {{DINOv2}: Learning Robust Visual Features without Supervision},
  year          = {2023},
  eprint        = {2304.07193},
  archivePrefix = {arXiv},
  primaryClass  = {cs.CV},
  doi           = {10.48550/arXiv.2304.07193}
}

@inproceedings{park2025resilient,
  author    = {Park, Konyul and Kim, Yecheol and Kim, Daehun and Choi, Jun Won},
  title     = {Resilient Sensor Fusion Under Adverse Sensor Failures via Multi-Modal Expert Fusion},
  booktitle = {Proceedings of the IEEE/CVF Conference on Computer Vision and Pattern Recognition (CVPR)},
  pages     = {6720--6729},
  year      = {2025}
}

@misc{hu2021lora,
  author        = {Hu, Edward J. and Shen, Yelong and Wallis, Phillip and Allen-Zhu, Zeyuan and Li, Yuanzhi and Wang, Shean and Wang, Lu and Chen, Weizhu},
  title         = {{LoRA}: Low-Rank Adaptation of Large Language Models},
  year          = {2021},
  eprint        = {2106.09685},
  archivePrefix = {arXiv},
  primaryClass  = {cs.CL},
  doi           = {10.48550/arXiv.2106.09685}
}

@inproceedings{banerjee2005meteor,
  author    = {Banerjee, Satanjeev and Lavie, Alon},
  title     = {{METEOR}: An Automatic Metric for {MT} Evaluation with Improved Correlation with Human Judgments},
  booktitle = {Proceedings of the ACL Workshop on Intrinsic and Extrinsic Evaluation Measures for Machine Translation and/or Summarization},
  pages     = {65--72},
  publisher = {Association for Computational Linguistics},
  address   = {Ann Arbor, Michigan},
  year      = {2005},
  url       = {https://aclanthology.org/W05-0909/}
}

@misc{wang2025internvl35,
  author        = {Wang, Weiyun and Gao, Zhangwei and Gu, Lixin and others},
  title         = {{InternVL3.5}: Advancing Open-Source Multimodal Models in Versatility, Reasoning, and Efficiency},
  year          = {2025},
  eprint        = {2508.18265},
  archivePrefix = {arXiv},
  primaryClass  = {cs.CV}
}

@misc{yu2025minicpmv45,
  author        = {Yu, Tianyu and Wang, Zefan and Wang, Chongyi and others},
  title         = {{MiniCPM-V 4.5}: Cooking Efficient {MLLMs} via Architecture, Data, and Training Recipe},
  year          = {2025},
  eprint        = {2509.18154},
  archivePrefix = {arXiv},
  primaryClass  = {cs.CV}
}

@misc{qwen2026qwen36,
  author       = {{Qwen Team}},
  title        = {{Qwen3.6-35B-A3B}: Agentic Coding Power, Now Open to All},
  year         = {2026},
  month        = apr,
  howpublished = {Official model release},
  url          = {https://qwen.ai/blog?id=qwen3.6-35b-a3b},
  note         = {Accessed: 2026-08-24}
}

@misc{qwen2026qwen38,
  author       = {{Qwen Team}},
  title        = {{Qwen3.8-Max}: A New Bar for Coding and Cowork},
  year         = {2026},
  month        = aug,
  howpublished = {Official model release},
  url          = {https://qwen.ai/blog?id=qwen3.8},
  note         = {Accessed: 2026-08-24}
}

@misc{qwen2026qwen3827b,
  author       = {{Qwen Team}},
  title        = {{Qwen3.8-27B}},
  year         = {2026},
  month        = aug,
  howpublished = {Official model card},
  url          = {https://huggingface.co/Qwen/Qwen3.8-27B},
  note         = {Accessed: 2026-08-24}
}

@misc{moonshot2026kimik3,
  author        = {{Kimi Team}},
  title         = {{Kimi K3}: Open Frontier Intelligence},
  year          = {2026},
  eprint        = {2607.24653},
  archivePrefix = {arXiv}
}

\end{document}